\documentclass[10pt,twocolumn,letterpaper]{article}

\usepackage{iccv}              

\definecolor{iccvblue}{rgb}{0.21,0.49,0.74}
\usepackage[pagebackref,breaklinks,colorlinks,allcolors=iccvblue]{hyperref}
\usepackage{multirow}
\usepackage{algorithm}      
\usepackage{algpseudocode}  
\usepackage{pifont}    
\usepackage{booktabs}  
\usepackage[accsupp]{axessibility}

\def\paperID{5129} 
\def\confName{ICCV}
\def\confYear{2025}

\title{CoST: Efficient Collaborative Perception From Unified Spatiotemporal Perspective}

\author{
Zongheng Tang$^{1,2}$ \quad Yi Liu$^{2}$ \quad Yifan Sun$^{2}$ \quad Yulu Gao$^{1,2}$ \quad Jinyu Chen$^{2}$ \quad
Runsheng Xu$^{3}$\thanks{Work done during Ph.D.} \quad Si Liu$^{2}$\thanks{Corresponding author.} \\
$^{1}$Hangzhou International Innovation Institute, Beihang University \\
$^{2}$School of Artificial Intelligence, Beihang University \quad
$^{3}$University of California, Los Angeles \\
{\tt\small \{tzhhhh123,18373214,gyl97,chenjinyu,liusi\}@buaa.edu.cn} \\
{\tt\small sunyf15@tsinghua.org.cn, rxx3386@g.ucla.edu}
}

\begin{document}
\maketitle
\begin{abstract}
Collaborative perception shares information among different agents and helps solving problems that individual agents may face, \eg, occlusions and small sensing range. Prior methods usually separate the multi-agent fusion and multi-time fusion into two consecutive steps. In contrast, this paper proposes an efficient collaborative perception that aggregates the observations from different agents (space) and different times into a unified spatio-temporal space simultaneously. The unified spatio-temporal space brings two benefits, \ie, efficient feature transmission and superior feature fusion. 1) Efficient feature transmission: each static object yields a single observation in the spatial temporal space, and thus only requires transmission only once (whereas prior methods re-transmit all the object features multiple times). 2) superior feature fusion: merging the multi-agent and multi-time fusion into a unified  spatial-temporal aggregation enables a more holistic perspective, thereby enhancing perception performance in challenging scenarios. Consequently, our \textbf{Co}llaborative perception with \textbf{S}patio-temporal \textbf{T}ransformer (CoST) gains improvement in both efficiency and accuracy. Notably, CoST is not tied to any specific method and is compatible with a majority of previous methods, enhancing their accuracy while reducing the transmission bandwidth. Code will be available at \url{https://github.com/tzhhhh123/CoST}.
\end{abstract}    
\begin{figure*}
   \vspace{-4mm}
    \begin{center}
    \includegraphics[width=0.9\linewidth]{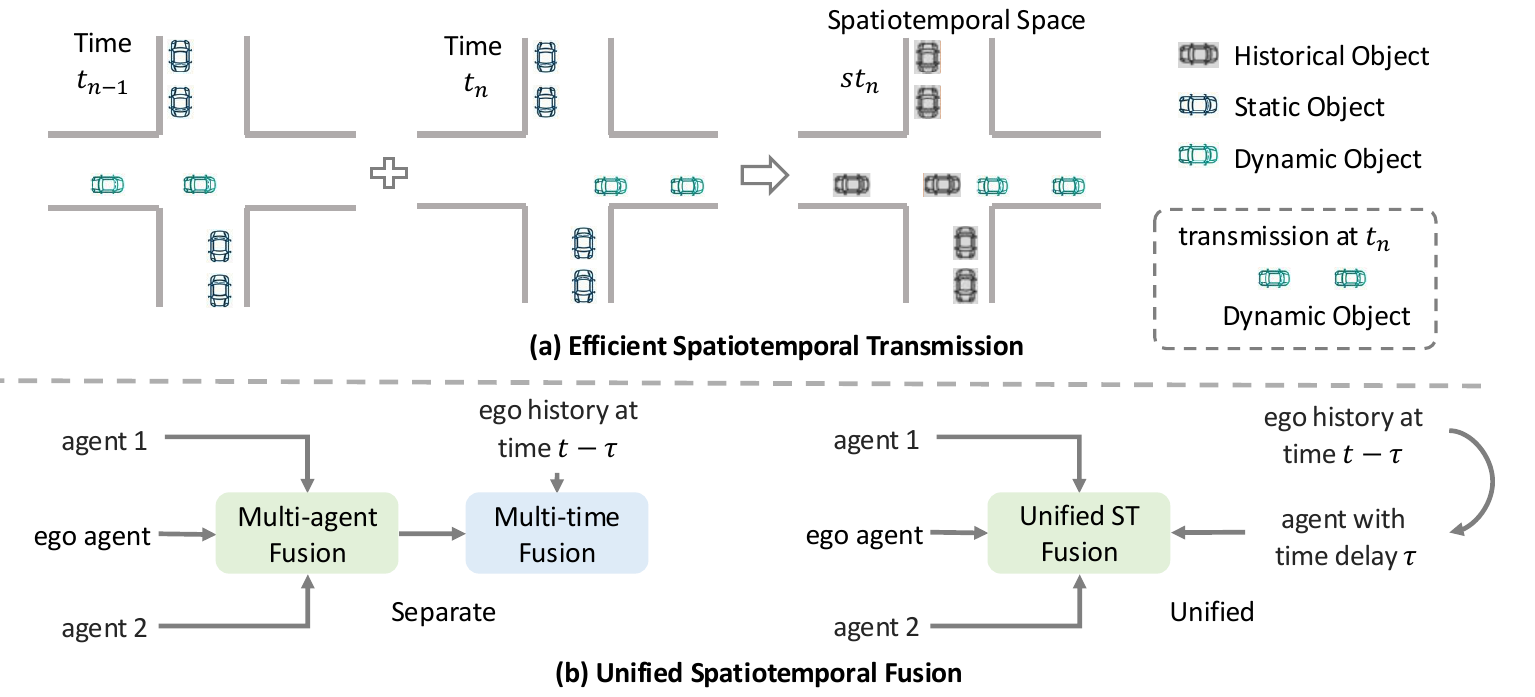}
    \end{center}
  \vspace{-6mm}
  \caption{
  (a) illustrates the core idea of our efficient Spatio-temporal Transmission (STT). By aggregating historical and current agents into a shared spatio-temporal space, static objects are retained from previous transmissions. Therefore, only dynamic object features need to be transmitted, significantly reducing bandwidth requirements.
  (b) presents our Unified Spatio-temporal Fusion (USTF) from spatio-temporal perspective. Contrary to traditional separate modeling methods, spatial and temporal fusion are unified in a single fusion module by viewing historical agents as duplicates of current agents (with a time delay). }
  \vspace{-4mm}
  \label{fig:motivation}
\end{figure*}

\section{Introduction}
\label{sec:intro}
Accurate 3D object detection is critical for autonomous driving to ensure road safety and traffic efficiency. Traditional single-agent systems, relying solely on onboard sensors, face limitations such as occlusions and restricted detection range, which pose safety risks. Multi-agent collaborative perception addresses these challenges by enabling agents to share complementary sensor data, fostering a holistic environmental understanding. In autonomous driving, collaborative perception facilitates information exchange between vehicles and infrastructure~\cite{wang2020v2vnet,xu2022opv2v,yu2022dair}, thereby overcoming single-agent constraints and enhancing perception accuracy and robustness.

A critical challenge in this domain is the integration of temporal information. Temporal data enriches the perception process by capturing past and future object states, correcting pose errors, and mitigating communication delays. Recent studies have underscored the importance of temporal modeling in this field. For instance, previous methods~\cite{lei2022latency, yang2024how2comm, yu2024flow} utilize historical data to enhance overall perception capabilities. While these works have made progress, they separate multi-agent and temporal fusion into sequential steps, processing either spatial or temporal data first. This separation prevents the full modeling of the interactions between spatial and temporal information, thereby limiting the overall capability of the perception system.

This paper proposes Collaborative Perception with Spatio-temporal Transformer (CoST), which utilizes a novel perspective to unify collaboration across space and time into a single spatio-temporal space.
Concretely, we view historical agents as duplicates of current agents (with some time delay), and they are directly comparable to each other. Therefore, given the current agents and historical agents, we can gather all their observations on objects into a shared spatio-temporal space for collaborative perception. This perspective no longer separates temporal modeling and multi-agent collaboration and is thus fundamentally different from previous temporal-modeling methods.

Our spatio-temporal perspective brings two important advantages, \emph{i.e.}, efficient feature transmission and superior feature fusion. The reasons are as follows:

$\bullet$ \emph{Efficient feature transmission.} 
In the spatio-temporal space, repeatedly transmitting static object features is redundant. Our Spatio-temporal Transmission (STT) module addresses this by transmitting only features of dynamic objects, while static ones are retrieved from prior frames via pose-projected reuse. As shown in Figure~\ref{fig:motivation}(a), among six objects in ${st}_{n}$, only two dynamic ones require transmission, substantially reducing communication bandwidth.

$\bullet$ \emph{Superior feature fusion.} 
Conventional methods treat multi-agent and temporal fusion as separate steps, often limiting interaction between ego history and other agents. In contrast, our Unified Spatio-temporal Fusion (USTF) views historical agents as time-delayed counterparts, enabling unified fusion across all agents and timestamps. As shown in Figure~\ref{fig:motivation}(b), this leads to a more comprehensive and consistent fusion process.

To mitigate the increased computational cost, USTF incorporates two strategies: (1) Recurrent modeling retains a single fused historical feature, lowering cost while maintaining long-term context; (2) A lightweight Multi-Agent Deformable Attention (MADA) module enables efficient cross-agent aggregation through sparse, learned attention sampling.

Although our specific implementation uses MADA, USTF itself is not tied to any specific multi-agent fusion method and serves as a general fusion approach. Additionally, the STT module can be used as a pluggable component, easily integrated into existing fusion methods to reduce transmission bandwidth.

\begin{figure*}[t]
    \vspace{-4mm}
    \begin{center}
    \includegraphics[width=1\linewidth]{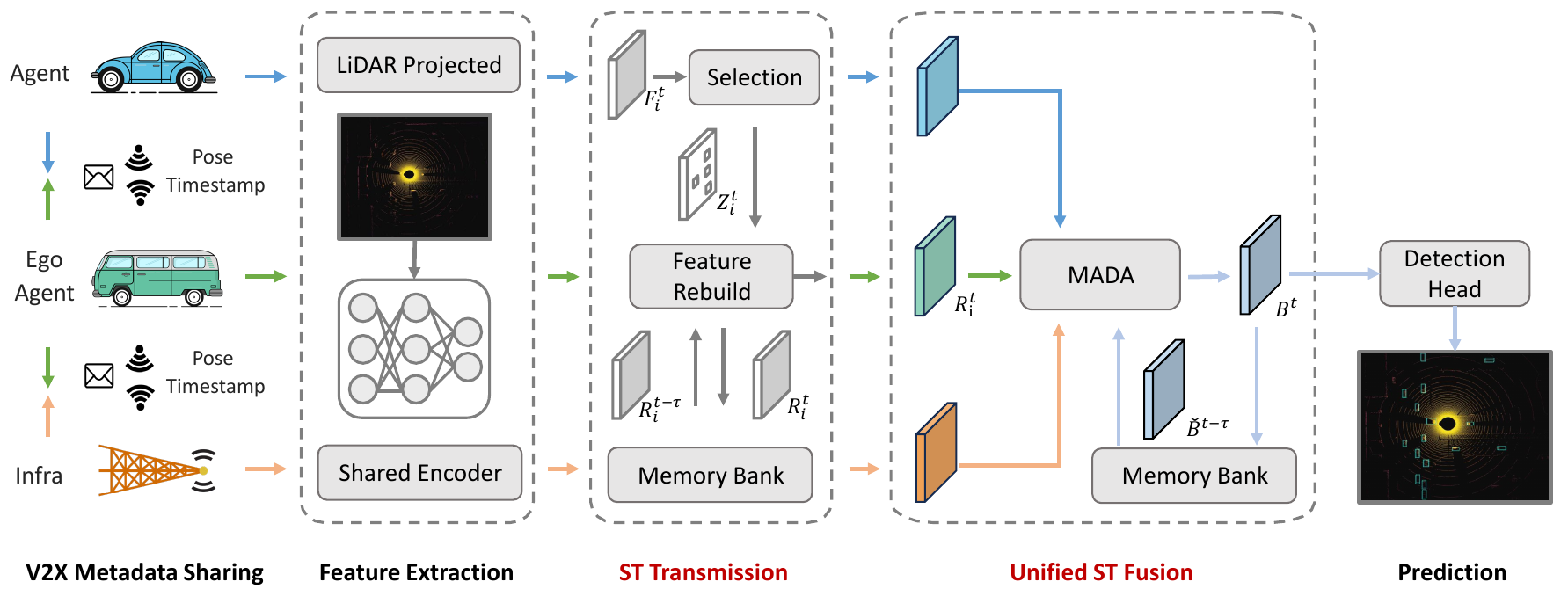}
    \end{center}
    \vspace{-4mm}
    \caption{Overview of the Cooperative Spatio-temporal Transformer (CoST) framework. CoST enhances feature transmission and fusion by integrating current and historical agents into a shared spatio-temporal space. The Spatio-temporal Transmission (STT) module selects and transmits only dynamic object features, significantly conserving bandwidth. These transmitted features are fused with historical features retrieved from the memory bank to reconstruct a comprehensive current feature. The Unified Spatio-temporal Fusion (USTF) module treats historical agents as delayed copies of current agents, consolidating multi-agent fusion into a single module, while the Multi-Agent Deformable Attention (MADA) module precisely aggregates features via learned offsets and weights. }
    \vspace{-4mm}
    \label{fig:framework}
\end{figure*}

CoST effectively enhances perception accuracy while reducing bandwidth, highlighting the crucial role of spatio-temporal dynamics.
In summary, the main contributions of this article can be summarized as follows:
\begin{itemize}
    \item We introduce the CoST framework to unify historical and current agents into a shared spatio-temporal space, reducing redundancy and enhancing accuracy.

    \item We introduce the STT module to selectively transmit dynamic object features, significantly reducing transmission bandwidth. Additionally, the USTF module integrates spatial and temporal fusion by treating historical agents as time-delayed duplicates of current agents, enabling more holistic and context-aware perception.

    \item We develop the MADA module as the specific implementation of the fusion module. Additionally, recurrent modeling is employed to reduce computational costs.
     
    \item Extensive experiments on V2XSet~\cite{xu2022v2x}, V2V4Real~\cite{xu2023v2v4real}, and DAIR-V2X~\cite{yu2022dair} demonstrate the superiority of our approach over existing methods. Additionally, our spatio-temporal-specific modules can seamlessly integrate with previous methods, enhancing accuracy and reducing bandwidth. The code will be available.
\end{itemize}

\section{Related Works}

\subsection{Collaborative Perception}
Collaborative perception has gained significant attention in multi-agent systems, particularly in autonomous vehicles (AVs), fueled by high-quality datasets~\cite{wang2020v2vnet,xu2022opv2v,xu2022v2x,xu2023v2v4real}. Fusion techniques in collaborative perception are typically categorized into early~\cite{chen2019cooper, arnold2020cooperative}, intermediate~\cite{liu2020who2com, li2021learning, xu2022bridging}, and late fusion~\cite{zeng2020dsdnet, shi2022vips}. Among these, intermediate fusion is favored for its efficient use of bandwidth while maintaining high accuracy. Methods such as OPV2V~\cite{xu2022opv2v} and V2X-ViT~\cite{xu2022v2x} advance fusion through graph-based and heterogeneous multi-agent attention mechanisms, respectively. In addition to fusion, minimizing transmission bandwidth while preserving accuracy remains a significant challenge. Approaches like Who2Com~\cite{liu2020who2com} and Where2comm~\cite{huwhere2comm} address bandwidth constraints by optimizing communicator selection and transmission content. Recent methods such as MRCNet~\cite{hong2023mrcnet} propose motion-aware communication to enhance robustness against pose errors and transmission delays. UMC~\cite{wang2023umc} introduces a unified and multi-resolution framework for bandwidth-efficient collaboration. Select2Col~\cite{liu2024select2col} further improves efficiency by leveraging spatio-temporal importance of semantic content for selective transmission. However, despite these advances, existing methods typically treat spatial and temporal aspects separately. In this work, we propose to seamlessly integrate spatial and temporal data in a unified spatio-temporal process.

\subsection{Temporal Modeling}
Temporal modeling has gained significant attention in 3D object detection~\cite{li2022bevformer, huang2022bevdet4d, li2022bevdepth, park2022time}, particularly in methods using Bird's Eye View (BEV). Works like BEVFormer~\cite{li2022bevformer}, BEVDet4D~\cite{huang2022bevdet4d}, and BEVDepth~\cite{li2022bevdepth} focus on warping BEV features from historical to current frames for efficient temporal modeling. Perspective-based methods, such as those based on DETR~\cite{carion2020detr}, use sparse queries to model object movement, as seen in Sparse4D~\cite{lin2022sparse4d} and Stream-PETR~\cite{wang2023exploring}, where sparse queries act as hidden states for temporal propagation. In collaborative perception for 3D detection, temporal data integration has become increasingly important. Methods like SyncNet~\cite{lei2022latency} and FFNet~\cite{yu2024flow} use historical data to predict future states and improve perception accuracy, while How2comm~\cite{yang2024how2comm} leverages temporal semantics to enhance contextual understanding and system robustness. Despite these advancements, current temporal modeling works have not effectively integrated temporal information with the unique characteristics of collaborative perception. In contrast, we propose treating historical agents as time-delayed duplicates of current agents, enabling a unified spatio-temporal fusion process.

\section{Methodology}
Under the spatio-temporal perspective, CoST offers an optimized transmission process and a unified fusion strategy. 

\subsection{Overview}
As depicted in Figure~\ref{fig:framework}, the collaborative perception process begins with V2X Metadata Sharing. In this stage, agents exchange meta-information, including 6 degrees of freedom (6DoF) pose, extrinsic parameters, velocity, and agent type. This meta-information serves as the foundation for the subsequent feature extraction and fusion stages.
Next, each agent performs Feature Extraction. An efficient method such as the PointPillar~\cite{lang2019pointpillars} is employed to transform sparse LiDAR point clouds into dense pillar tensors. These tensors are then processed to generate semantic Bird's Eye View (BEV) features, denoted as $\mathbf{F}_i^{t} \in \mathbb{R}^{H \times W \times C}$, at timestamp $t$ for agent $i$. 
After feature extraction, agents proceed to the Spatio-temporal Transmission (STT) stage. In this phase, agents exchange features; the STT module ensures that only dynamic object features are transmitted while static object features are retrieved from the memory bank. The ego agent fuses each received dynamic feature with its corresponding historical feature, reconstructing a comprehensive feature $\mathbf{R}_i^t \in \mathbb{R}^{H \times W \times C}$, which is then stored in memory for future use. 
Following this, agents move to the Unified Spatio-temporal Fusion (USTF) stage, where the ego agent merges its own features with those received from other agents to form a comprehensive collaborative feature $\mathbf{B} \in \mathbb{R}^{H \times W \times C}$ at the current timestamp. 
Finally, the fused perceptual features are passed to the Detection Head to predict the 3D object locations, dimensions, and classifications. The detection head consists of two $1 \times 1$ convolution layers: one for box regression, which predicts position, dimensions, and yaw angle, and another for classification, which outputs a confidence score. The loss functions include smooth $L1$ loss~\cite{ren2016faster} for regression and focal loss~\cite{lin2017focal} for classification. 
More details are provided in supplementary material.

\subsection{Spatio-temporal Transmission}  
The STT module in our CoST framework reduces communication bandwidth by transmitting only dynamic object features instead of the full BEV map. Conventional methods typically compress feature channels~\cite{xu2022v2x,xu2023v2v4real}, but neglect temporal redundancy where static regions remain unchanged. While Where2comm~\cite{huwhere2comm} transmits sparse tokens from salient regions, it does not leverage the fact that static areas can be reconstructed from prior frames. Our method explicitly focuses on transmitting dynamic content, avoiding redundant communication while preserving scene completeness. To identify dynamic regions, we first compute the saliency map:
\begin{equation}
E_i^{t} = \operatorname{Sigmoid}(\phi(\mathbf{F}_i^{t})) \in [0,1]^{H \times W},
\end{equation}
where $\phi(\cdot)$ denotes the classification head shared with the detection module.

We then calculate the dynamic map to capture temporal changes:
\begin{equation}
D_i^{t} = \left| E_i^{t} - E_i^{t-\tau} \right|,
\end{equation}
where $\tau$ is the temporal gap between current frame $t$ and the last transmission. This step highlights newly appeared or disappeared objects.

Next, a binary selection mask $M_i^{t} \in \{0,1\}^{H \times W}$ is computed by blending the dynamic and saliency maps, emphasizing dynamic and high-importance regions. The selected feature tokens are obtained via:
\begin{equation}
z_i^{t} = M_i^{t} \odot \mathbf{F}_i^{t},
\end{equation}
where $\odot$ denotes element-wise multiplication.

The ego agent (indexed as $1$) receives the transmitted $z_i^t$ from other agents ($i \neq 1$). These sparse features are aligned to the ego frame using relative poses via affine warping and bilinear resampling. They are then combined with historical features $\mathbf{R}_i^{t - \tau}$ using a reconstruction module (consisting of convolutional layers, normalization, and GELU activation) to produce the reconstructed feature $\mathbf{R}_i^t$.

Algorithm~\ref{alg:tc} summarizes this spatio-temporal compression and reconstruction process. By transmitting only dynamic and salient features and leveraging temporal alignment, CoST achieves strong reconstruction quality with significantly reduced communication cost. An overview illustration is provided in the supplementary material.

Notably, STT can be seamlessly integrated into existing multi-agent fusion frameworks by replacing the original compression module (e.g., downsampling/upsampling convolutions), without affecting the downstream fusion design.

\begin{algorithm}[t]
    \caption{STT With {\color{blue}Saliency} and {\color{orange}Dynamic} Map}
    \label{alg:tc}
    \begin{tabular}{l@{}l}
        \textbf{Input: } 
        & current feature $\{F^{t}\}_{i=1}^N$, rescale rate $\rho$,\\
        & historical feature $\{R^{t-\tau}\}_{i=1}^N$, temporal interval $\tau$.
    \end{tabular}
    
    \begin{tabular}{l@{}l}
        \textbf{Output: } & reconstructed feature $\{R^{t}\}_{i=1}^N$.
    \end{tabular}
    
    \begin{algorithmic}[1]
        \For{$i=2$ \text{to} $N$}
            \State  ${\color{blue}E_i^{t}} = Sigmoid(\phi(\mathbf{F}_i^{t}))$ 
            
            \State  ${\color{blue}E_i^{t-\tau}} = Sigmoid(\phi(\mathbf{R}_i^{t-\tau}))$ 
            
            \State  ${\color{orange}D_i^{t}} = |{\color{blue}E_i^{t}} - {\color{blue}E_i^{t-\tau}}|$ 
            
            \State  $M_i^{t} = {\color{blue}E_i^{t}} \cdot \left( \frac{1}{\rho + 1} + {\color{orange}D_i^{t}} \cdot \rho \right)$  
            
            
            \State  $z_i^{t} = \mathbf{F}_i^{t} \odot M_i^{t}$  
    
            \State transfer $z_i^{t}$ from agent $i$ to ego agent $1$
            \State receive $z_i^{t}$ at ego agent $1$ from agent $i$
            \State reconstruct $\mathbf{R}_i^{t}$ with $z_i^{t}$ and $\mathbf{R}_i^{t-\tau}$ 
        \EndFor
    \end{algorithmic}
\end{algorithm}

\begin{figure*}[ht]
    \begin{center}
    \includegraphics[width=1\linewidth]{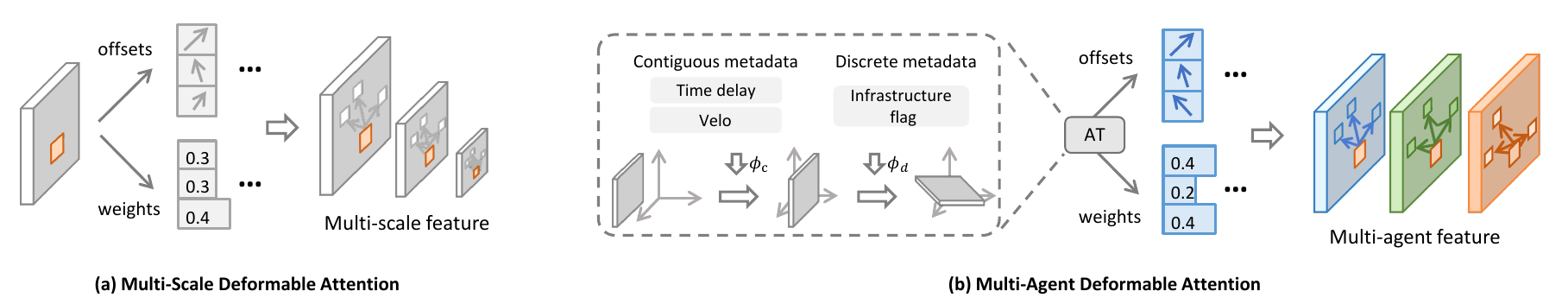}
    \end{center}
    \vspace{-4mm}
    \caption{(a) illustrates the Multi-Scale Deformable Attention mechanism, where deformable operations highlight relevant regions across various feature scales. (b) showcases our advanced Multi-Agent Deformable Attention mechanism. It starts with integrating agent-specific attributes, such as latency and velocity, through an Align Transformation (AT) to refine the features. Subsequently, deformable attention for each agent's features is computed, incorporating the ego features from other agents to enhance the collective perception.}
    \label{fig:fig4}
\end{figure*}

\begin{figure}[t]
    \centering
    \includegraphics[width=1\linewidth]{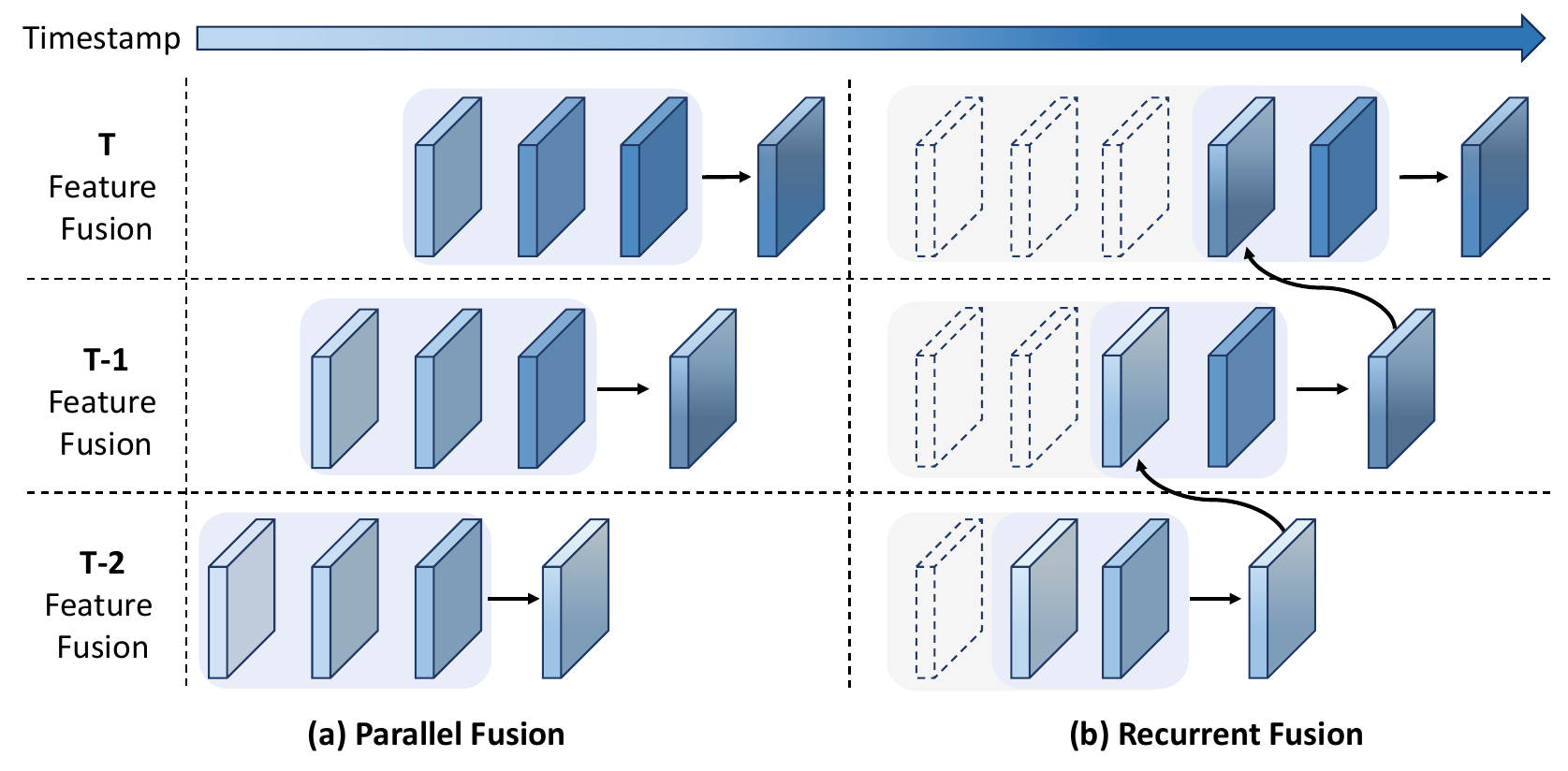}
    \vspace{-5mm}
    \caption{This diagram contrasts parallel fusion (a) with recurrent fusion (b) in the context of historical feature integration. }
    \vspace{-3mm}
    \label{fig:recurrent}
\end{figure}

\subsection{Unified Spatio-temporal Fusion}
The second module in our CoST framework is the Unified Spatio-temporal Fusion (USTF) module, which integrates spatial and temporal features to enhance perception accuracy. Unlike previous methods that handle multi-agent and temporal fusion separately, USTF treats historical agents as time-delayed duplicates of current agents. By tagging historical agents at time $t-\tau$ with a fixed delay $\tau$, they become equivalent to current agents and can be fused within a single multi-agent fusion module.

Directly incorporating all historical agents increases computational cost. As shown in Figure~\ref{fig:recurrent}(a), previous parallel fusion methods aggregate features from a fixed-length window of size $L$, leading to $(N+1)L$ agents (with $N$ agents per frame). In contrast, our recurrent modeling (Figure~\ref{fig:recurrent}(b)) retains only one fused historical feature, reducing the total to $N+1$ agents while capturing longer-term context beyond the window size.

Specifically, let $\mathbf{B}^{t-\tau}$ denote the fused collaborative feature at time $t-\tau$. We project $\mathbf{B}^{t-\tau}$ to $\check{\mathbf{B}}^{t-\tau}$ using the relative pose matrices between $t-\tau$ and $t$, so that $\check{\mathbf{B}}^{t-\tau}$ can be treated as an independent agent with delay $\tau$. This delayed agent is then merged with current agent features. 

USTF is fusion-agnostic and compatible with any multi-agent fusion architecture. The previously fused collaborative feature $\mathbf{B}^{t-\tau}$ is projected to the current frame using relative pose matrices, resulting in $\check{\mathbf{B}}^{t-\tau}$, which is treated as an additional agent with a fixed delay. This leads to $(N{+}1)$ input agents to the fusion module, enabling unified spatio-temporal modeling while maintaining compatibility with the original pipeline.

\subsection{Multi-Agent Deformable Attention and Align Transformation}
\label{sec:mada}
The third module in our CoST framework is the Multi-Agent Deformable Attention with Align Transformation (MADA), which efficiently aggregates features across agents. While recent studies employ attention mechanisms for multi-agent fusion~\cite{xu2022v2x}, these methods can be computationally expensive. Inspired by deformable attention~\cite{zhu2020deformable}, our MADA module improves efficiency by selectively sampling informative regions across agents. As illustrated in Figure~\ref{fig:fig4}, our approach first injects agent-specific meta information via the Align Transformation (AT) module and then aggregates features across agents.

To enhance agent interactions, we inject agent-specific meta information (e.g., speed, latency, infrastructure markers) using the AT module. Denote the multi-agent feature set as $V = \{\mathbf{R}_i^{t}\}_{i=1}^N$, and the corresponding discrete and continuous meta information as $S^{t} = \{S_i^{t}\}_{i=1}^N$ and $U^{t} = \{U_i^{t}\}_{i=1}^N$. We embed this information via learnable projections:
\begin{equation}
V' = \operatorname{AT}(V, S^{t}, U^{t}) = \varphi_c(S^{t})V + \varphi_d(U^{t}),
\end{equation}
where $\varphi_c$ and $\varphi_d$ are learnable layers, yielding meta-embedded features $V'$.

Let $V' = \{x^i\}_{i=1}^N$, with $x^i \in \mathbb{R}^{C \times H \times W}$. For a query with feature $z_q$ and 2D reference point $p_q$, MADA is defined as:
\begin{equation}
\begin{split}
& \operatorname{MADA}(z_q, p_q, \{x^i\}_{i=1}^N) = \\
& \sum^{M}_{m=1} W_m\left[ \sum_{i=1}^{N}\sum_{k=1}^K A_{miqk} \cdot W'_m x^{i}(p_q+\Delta p_{miqk}) \right],
\end{split}
\end{equation}
where $m$ indexes the $M$ attention heads, $k$ indexes the $K$ sampling points, and $W_m$, $W'_m$ are learnable weights. The query feature $z_q$ is projected to generate sampling offsets $\Delta p_{miqk}$ and attention weights $A_{miqk}$ for each sampling point. The aggregated result is then processed by a feed-forward network and a mean operation to yield the final features for the detection head.

\renewcommand\arraystretch{1.3}
    {\fontsize{3.2}{3.5}\selectfont
    \setlength{\tabcolsep}{7.5pt}
    \setlength{\arrayrulewidth}{0.3pt}
\begin{table*}[]
\centering
\label{tab:main}
\renewcommand{\arraystretch}{0.9}
\begin{tabular}{@{}cccccccccc@{}}
\toprule
\multirow{2}{*}{Dataset}           & \multirow{2}{*}{Methods}         & \multicolumn{4}{c}{AP@0.5}               & \multicolumn{4}{c}{AP@0.7}              \\ \cmidrule(l){3-10} 
               &        & Overall         & Long      & Middle    & \multicolumn{1}{c|}{Short}     & Overall         & Long      & Middle    & Short     \\ \midrule

\multicolumn{1}{c|}{\multirow{6}{*}{V2V4Real}} & \multicolumn{1}{c|}{Early Fusion}      & 0.587    & 0.464    & 0.409    & \multicolumn{1}{c|}{0.769}    & 0.286    & 0.207     & 0.196    & 0.399    \\
\multicolumn{1}{c|}{}      & \multicolumn{1}{c|}{Late Fusion}   & 0.550     & 0.363     & 0.437     & \multicolumn{1}{c|}{0.734}     & 0.266    & 0.173    & 0.222    & 0.366    \\
\multicolumn{1}{c|}{}      & \multicolumn{1}{c|}{No Fusion}     & 0.416     & 0.089     & 0.298     & \multicolumn{1}{c|}{0.655}     & 0.218    & 0.039    & 0.167    & 0.346    \\
\multicolumn{1}{c|}{}      & \multicolumn{1}{c|}{F-Cooper~\cite{fcooper}} & 0.645 & 0.459 & 0.511 & \multicolumn{1}{c|}{0.798}     & 0.354    & 0.178    & 0.275    & 0.503    \\
\multicolumn{1}{c|}{}      & \multicolumn{1}{c|}{V2X-ViT~\cite{xu2022v2x}}      & \underline{0.662} & 0.473    & 0.552     & \multicolumn{1}{c|}{0.803}     & 0.365    & 0.171     & 0.314    & 0.517     \\
\multicolumn{1}{c|}{}      & \multicolumn{1}{c|}{OPV2V~\cite{xu2022opv2v}}       & 0.654    & 0.456    & 0.520    & \multicolumn{1}{c|}{0.812}    & 0.375    & 0.188    & 0.320    & 0.516    \\
\multicolumn{1}{c|}{}      & \multicolumn{1}{c|}{CoBEVT~\cite{xu2022cobevt}}      & 0.656      & 0.461    & 0.500     & \multicolumn{1}{c|}{0.823}    & \underline{0.379}    & 0.187    & 0.280    & 0.561    \\
\multicolumn{1}{c|}{}      & \multicolumn{1}{c|}{\textbf{CoST}}        & \textbf{0.710} & 0.521 & 0.598 & \multicolumn{1}{c|}{0.855} & \textbf{0.440} & 0.263 & 0.354 & 0.589 \\ \midrule
\hline
\multicolumn{1}{c|}{\multirow{8}{*}{V2XSet}}   & \multicolumn{1}{c|}{Early Fusion}      & 0.842    & 0.717    & 0.882    & \multicolumn{1}{c|}{0.957}    & 0.741     & 0.554    & 0.793    & 0.920      \\
\multicolumn{1}{c|}{}      & \multicolumn{1}{c|}{Late Fusion} & 0.741       & 0.526         & 0.764     & \multicolumn{1}{c|}{0.915}     & 0.634    & 0.387    & 0.646    & 0.852    \\
\multicolumn{1}{c|}{}      & \multicolumn{1}{c|}{No Fusion} & 0.613         & 0.362         & 0.687     & \multicolumn{1}{c|}{0.810}     & 0.447    & 0.169    & 0.488    & 0.718    \\
\multicolumn{1}{c|}{}      & \multicolumn{1}{c|}{DiscoNet~\cite{mehr2019disconet}}    & 0.863    & 0.746    & 0.894    & \multicolumn{1}{c|}{0.957}    & 0.761    & 0.564    & 0.808     & 0.926    \\
\multicolumn{1}{c|}{}      & \multicolumn{1}{c|}{V2VNet~\cite{wang2020v2vnet}}      & \underline{0.881}    & 0.779    & 0.911    & \multicolumn{1}{c|}{0.972}    & 0.754    & 0.559    & 0.814    & 0.925    \\
\multicolumn{1}{c|}{}      & \multicolumn{1}{c|}{V2X-ViT~\cite{xu2022v2x}}      & 0.872    & 0.747    & 0.886    & \multicolumn{1}{c|}{0.962}    & 0.756    & 0.537     & 0.787    & 0.928    \\
\multicolumn{1}{c|}{}      & \multicolumn{1}{c|}{Where2comm~\cite{huwhere2comm}}        & 0.823    & 0.664    & 0.870    & \multicolumn{1}{c|}{0.943}    & 0.706    & 0.474    & 0.751    & 0.894    \\
\multicolumn{1}{c|}{}      & \multicolumn{1}{c|}{OPV2V~\cite{xu2022opv2v}}       & 0.856     & 0.749    & 0.894    & \multicolumn{1}{c|}{0.949}    & 0.751     & 0.561    & 0.811    & 0.915     \\
\multicolumn{1}{c|}{}      & \multicolumn{1}{c|}{CoBEVT~\cite{xu2022cobevt}}      & 0.876    & 0.765    & 0.896    & \multicolumn{1}{c|}{0.967}    & \underline{0.771}    & 0.566    & 0.815    & 0.938    \\
\multicolumn{1}{c|}{}      & \multicolumn{1}{c|}{\textbf{CoST}}         & \textbf{0.915} & 0.794 & 0.910  & \multicolumn{1}{c|}{0.975}          & \textbf{0.814} & 0.595  & 0.816 & 0.950 \\ \midrule
\hline
\multicolumn{1}{c|}{\multirow{8}{*}{DAIR-V2X}} & \multicolumn{1}{c|}{Early Fusion}      & \underline{0.608}    & 0.477    & 0.621     & \multicolumn{1}{c|}{0.746}    & 0.434    & 0.311    & 0.449    & 0.585     \\
\multicolumn{1}{c|}{}      & \multicolumn{1}{c|}{Late Fusion} & 0.568       & 0.530         & 0.546     & \multicolumn{1}{c|}{0.634}     & 0.403    & 0.360    & 0.387    & 0.473    \\
\multicolumn{1}{c|}{}      & \multicolumn{1}{c|}{No Fusion} & 0.540         & 0.389         & 0.583     & \multicolumn{1}{c|}{0.665}     & \underline{0.466}    & 0.316    & 0.500    & 0.609    \\
\multicolumn{1}{c|}{}      & \multicolumn{1}{c|}{DiscoNet~\cite{mehr2019disconet}}    & 0.546    & 0.370      & 0.614    & \multicolumn{1}{c|}{0.709}    & 0.418    & 0.259    & 0.484    & 0.581    \\
\multicolumn{1}{c|}{}      & \multicolumn{1}{c|}{V2VNet~\cite{wang2020v2vnet}}      & 0.578     & 0.452    & 0.629    & \multicolumn{1}{c|}{0.735}    & 0.444    & 0.327    & 0.486    & 0.613    \\
\multicolumn{1}{c|}{}      & \multicolumn{1}{c|}{V2X-ViT~\cite{xu2022v2x}}      & 0.583    & 0.451    & 0.639    & \multicolumn{1}{c|}{0.697}    & 0.439    & 0.311    & 0.479     & 0.575    \\
\multicolumn{1}{c|}{}      & \multicolumn{1}{c|}{Where2comm~\cite{huwhere2comm}}        & 0.591    & 0.482 & 0.651 & \multicolumn{1}{c|}{0.731}    & 0.460    & 0.348 & 0.508 & 0.629     \\
\multicolumn{1}{c|}{}      & \multicolumn{1}{c|}{OPV2V~\cite{xu2022opv2v}}       & 0.556    & 0.434    & 0.607    & \multicolumn{1}{c|}{0.708}    & 0.388    & 0.278    & 0.417    & 0.560    \\
\multicolumn{1}{c|}{}      & \multicolumn{1}{c|}{CoBEVT~\cite{xu2022cobevt}}      & 0.580    & 0.458    & 0.625    & \multicolumn{1}{c|}{0.734}    & 0.433    & 0.307    & 0.473    & 0.618    \\
\multicolumn{1}{c|}{}      & \multicolumn{1}{c|}{TransIFF$\dagger$~\cite{chen2023transiff}}       & 0.596    & -    & -    & \multicolumn{1}{c|}{-}    & 0.460    & -    & -    & -    \\
\multicolumn{1}{c|}{}      & \multicolumn{1}{c|}{What2comm$\dagger$~\cite{yang2023what2comm}}       & \underline{0.608}    & -    & -    & \multicolumn{1}{c|}{-}    & 0.446    & -    & -    & -    \\
\multicolumn{1}{c|}{}      & \multicolumn{1}{c|}{\textbf{CoST}}        & \textbf{0.614} & 0.477    & 0.637    & \multicolumn{1}{c|}{0.759} & \textbf{0.473} & 0.329    & 0.498    & 0.640 \\ \bottomrule
\end{tabular}
\caption{
Summary and comparison of essential characteristics for different methods on V2V4Real, DAIR-V2X, and V2XSet datasets. The symbol $\dagger$ indicates that the results from the original paper include only overall metrics. \textbf{Best} and \underline{second-best} results are highlighted.
}

\label{tab:main}
\end{table*}
}
\section{Experiments}
\subsection{Datasets}
To comprehensively evaluate the performance of CoST,  we conduct experiments on three well-known datasets: V2V4Real~\cite{xu2023v2v4real}, V2XSet~\cite{xu2022v2x}, and DAIR-V2X~\cite{yu2022dair}.
V2V4Real offers a wide range of real-world V2V communication scenarios, V2XSet combines V2V and V2I interactions with simulated noise, and DAIR-V2X focuses on real-world V2I communication, covering various collaborative perception scenarios. 
Dataset and training details are provided in the supplementary material.

\subsection{Experimental setup}

\noindent {\bf{Evaluation Metric:}}  
We evaluate LiDAR-based vehicle detection, counting any vehicle detected by connected agents. Performance is measured by Average Precision (AP) at IoU thresholds of 0.5 and 0.7, with AP reported for short (0-30m), middle (30-50m), and long (50-100m) ranges.

\noindent {\bf{Compared Methods:}}  
We compare our results with ~\cite{liu2023towards}. The baseline "No Fusion" uses only the vehicle's own LiDAR data. We also evaluate "Late Fusion", which combines detections from multiple agents using non-maximum suppression, and "Early Fusion", which merges raw LiDAR data from nearby agents. Furthermore, we compare advanced fusion techniques, including OPV2V~\cite{xu2022opv2v}, F-Cooper~\cite{fcooper}, V2VNet~\cite{wang2020v2vnet}, DiscoNet~\cite{mehr2019disconet}, COBEVT~\cite{xu2022cobevt}, and Where2Comm~\cite{huwhere2comm}, all employing the PointPillar~\cite{lang2019pointpillars} architecture for feature extraction.

\subsection{Main Results}
Assessing the accuracy of object detection at different distances plays a crucial role in 3D detection research, providing valuable insights into how detection performance changes with varying distances. 

Our study aimed to analyze object detection accuracy across specific distance ranges, namely 0-30m, 30-50m, and 50-100m, measured from the ego vehicle.

As depicted in Table \ref{tab:main}, our proposed CoST demonstrated superior performance on the V2V4Real, V2XSet, and DAIR-V2X datasets, surpassing existing methodologies by considerable margins. Notably, on the V2V4Real dataset, CoST enhanced the AP@0.5 and AP@0.7 scores by $4.8\%$ and $6.1\%$, respectively, compared to the previously reported best results. 
For the DAIR-V2X dataset, our method achieved an AP@0.5 of $61.4\%$ and an AP@0.7 of $47.3\%$. 
Moreover, on the V2XSet dataset, our method achieved an AP@0.5 of $91.5\%$ and an AP@0.7 of $81.4\%$, improving the previous SOTA by $3.4\%$ and $4.3\%$, respectively.

These advancements highlight the effectiveness of our spatio-temporal modeling, especially in scenarios demanding accurate long-range detection.

\begin{table}[]
\centering
\label{tab:main_abl}
\renewcommand{\arraystretch}{0.9}
\setlength{\tabcolsep}{4mm}{
\begin{tabular}{@{}ccccc@{}}
\toprule
Spatial      & AT        & Temporal       & AP@0.5 & AP@0.7 \\ \midrule
\ding{52} &           &           & 0.6681    & 0.4069    \\
\ding{52} & \ding{52} &           & 0.6768    & 0.4093    \\
\ding{52} &           & \ding{52} & 0.6901    & 0.4279    \\
\ding{52} & \ding{52} & \ding{52} & \textbf{0.7096}    & \textbf{0.4397}    \\ \bottomrule
\end{tabular}}
\vspace{-2mm}
\caption{Component ablation study of MADA, Spatial, AT,
Temporal represents i) only current agents, ii) adding Align Transformation, and iii) adding historical agents, respectively}
\vspace{-1mm}
\label{tab:main_abl}
\end{table}
\begin{table}[thbp]
\centering
\renewcommand{\arraystretch}{0.9}
\setlength{\tabcolsep}{3.5mm}{
\begin{tabular}{@{}cccc@{}}
\toprule
Input Frames & Time Interval & AP@0.5 & AP@0.7 \\ \midrule
1 & - & 0.6768 & 0.4093 \\ \midrule
\multirow{2}{*}{2} & 1 & 0.6944 & 0.4292 \\
 & 2 & 0.7064 & 0.4361 \\ \midrule
\multirow{2}{*}{3} & 1 & 0.7059 & 0.4301 \\
 & 2 & 0.7096 & 0.4397 \\ \bottomrule
\end{tabular}}
\vspace{-2mm}
\caption{Ablation study on the impact of input frames and time intervals(frames) on model performance.}
\vspace{-2mm}
\label{table:temporal_frame_ablation}
\end{table}

\subsection{Ablation Study}
This section aims to evaluate the impact of specific components within the CoST architecture. 

\textbf{Component Analysis.}

We analyze each component as shown in Table~\ref{tab:main_abl}.

Incorporating only MADA yields noticeable improvement, achieving $66.81\%$ AP@0.5 and $40.69\%$ AP@0.7.
The introduction of AT results in an increase of $0.87\%$ AP@0.5.
This enhancement can be attributed to AT's capability to refine the localization of detected objects, aligning them more accurately with the ground truth.
Moreover, the integration of USTF demonstrates the most substantial improvements, elevating AP@0.5 and AP@0.7 to $69.01\%$ and $42.79\%$, respectively. 
Finally, the combined utilization of Align Transformation and Recurrent Temporal Modeling leads to a further accuracy improvement of $4.15\%$ and $3.28\%$ for AP@0.5 and AP@0.7.

\textbf{Historical Frames and Interval.} The ablation study on the impact of input frames and time intervals on the V2V4Real Dataset, as shown in Table \ref{table:temporal_frame_ablation}, reveals their significant influence on model performance. 
Using a single input frame, the model achieves an AP@0.5 of 67.68\% and an AP@0.7 of 40.93\%. 
Introducing a second frame with a time interval of 1 improves the performance to an AP@0.5 of 69.44\% and an AP@0.7 of 42.92\%. 
Increasing the interval to 2 provides the highest performance, with an AP@0.5 of 70.96\% and an AP@0.7 of 43.97\%.
These results confirm the benefit of using multiple frames and appropriate intervals for accuracy and robustness.

\begin{figure}[tbp]
    \centering
    \includegraphics[width=0.9\linewidth]{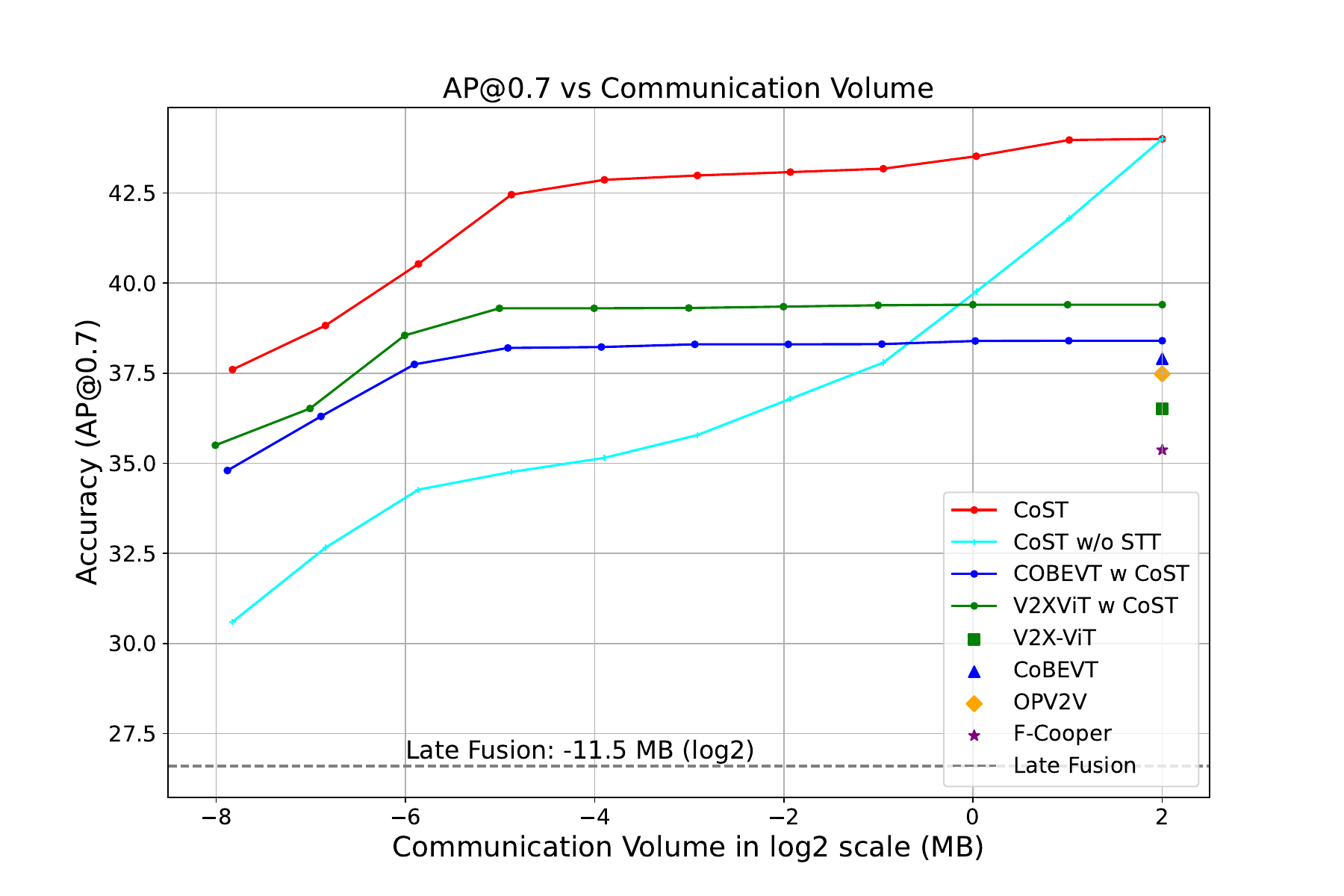}
    \vspace{-3mm}
    \caption{Performance-Bandwidth Trade-off. The x-axis represents the Bandwidth in log2 scale, and the y-axis represents the AP@0.7 performance on the V2V4Real dataset.}
    \vspace{-3mm}
    \label{fig:compress}
\end{figure}

\begin{figure*}[htbp]
    \centering
    \captionsetup[subfloat]{font=scriptsize}
    \subfloat[Localization Error]{
        \label{positional_error_dair}
        \label{heading_error_dair}
        \label{error_dair}
        \includegraphics[height=0.15\textheight,trim=12 5 5 10,clip]{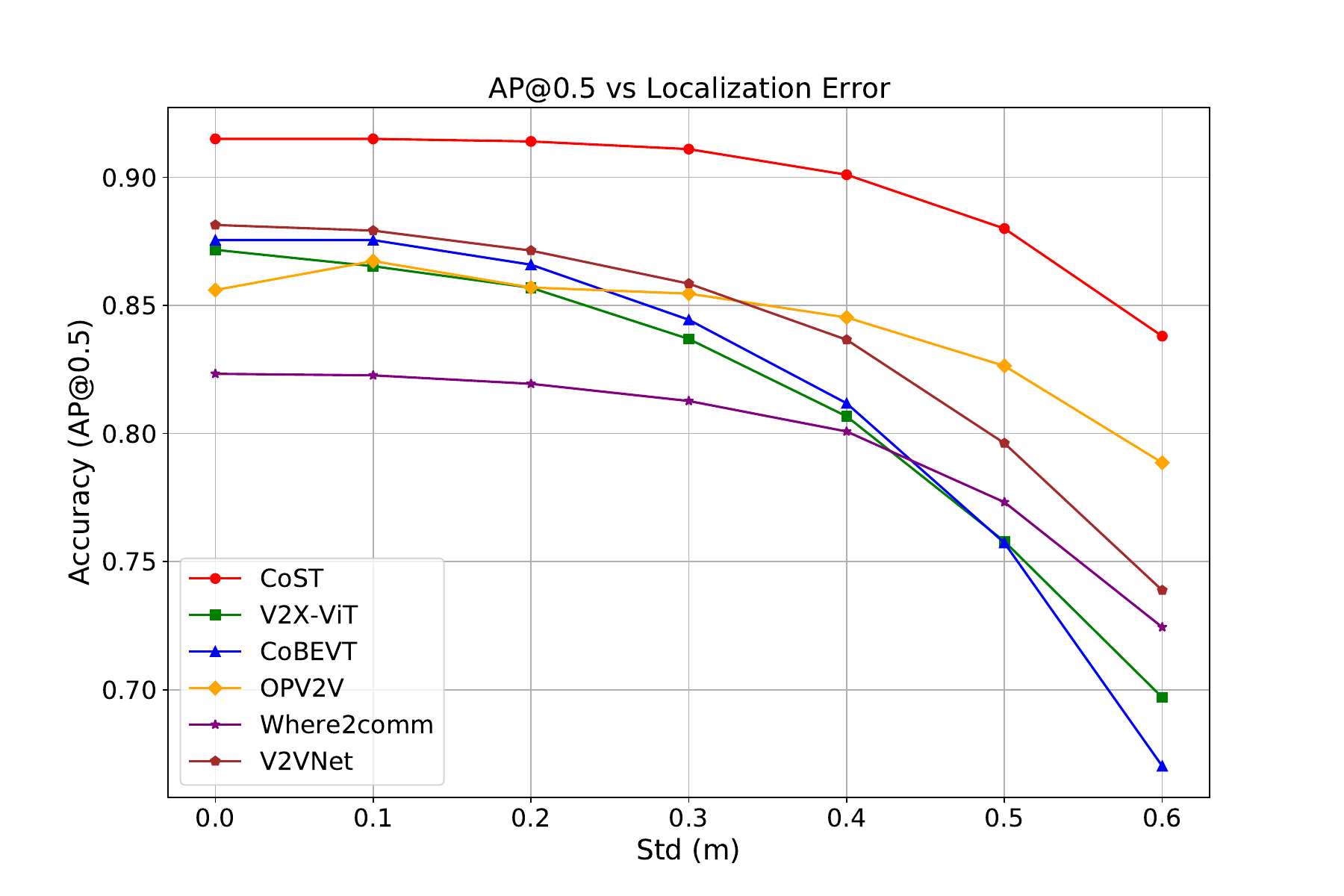}
    }
    \subfloat[Rotation Error]{
        \label{positional_error_v2xset}
        \label{heading_error_v2xset}
        \label{error_v2xset}
        \includegraphics[height=0.15\textheight,trim=12 5 5 10,clip]{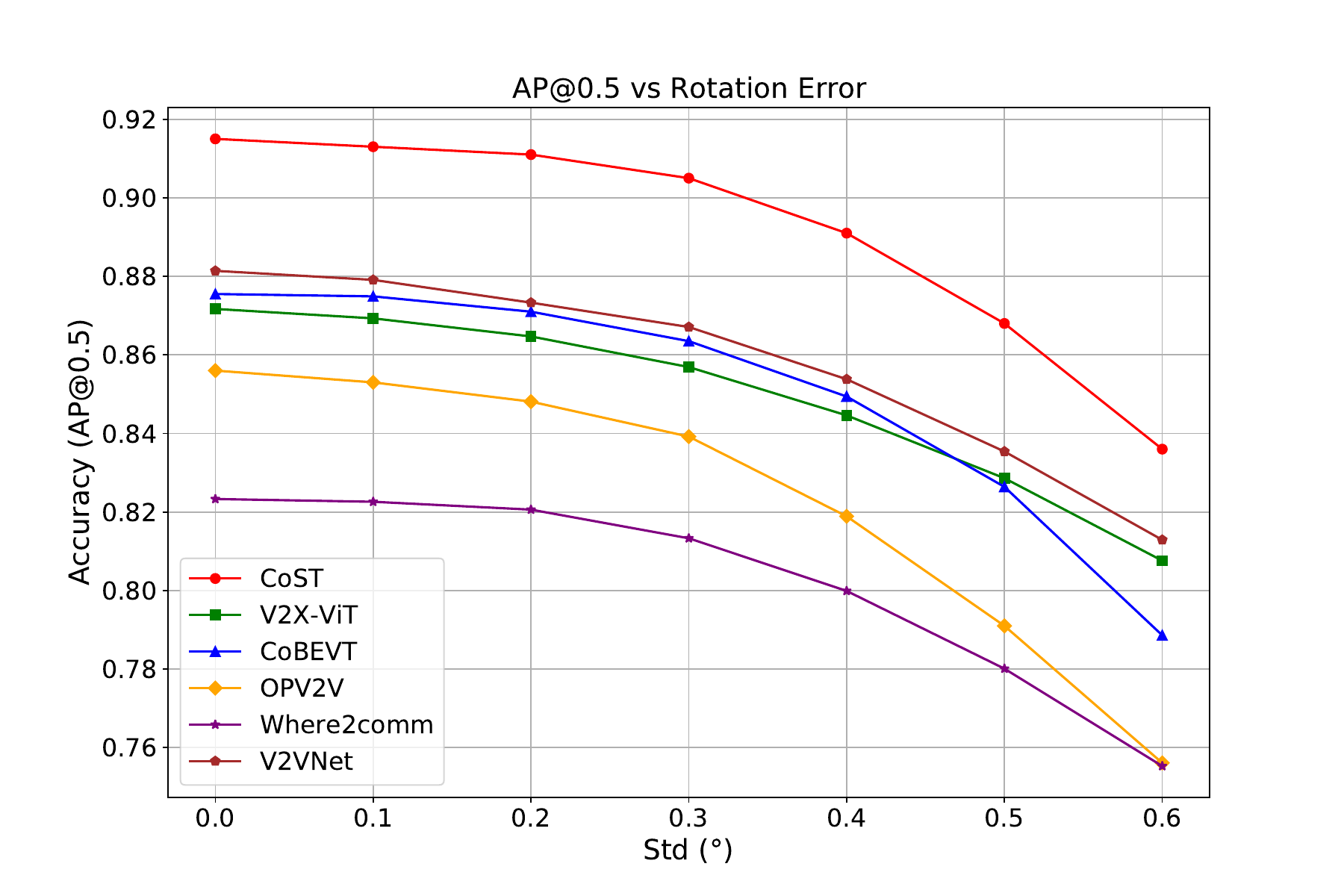} 
    }
    \subfloat[Transmission Latency]{
        \label{positional_error_v2xset}
        \label{heading_error_v2xset}
        \label{error_v2xset}
        \includegraphics[height=0.15\textheight,trim=12 5 5 10,clip]{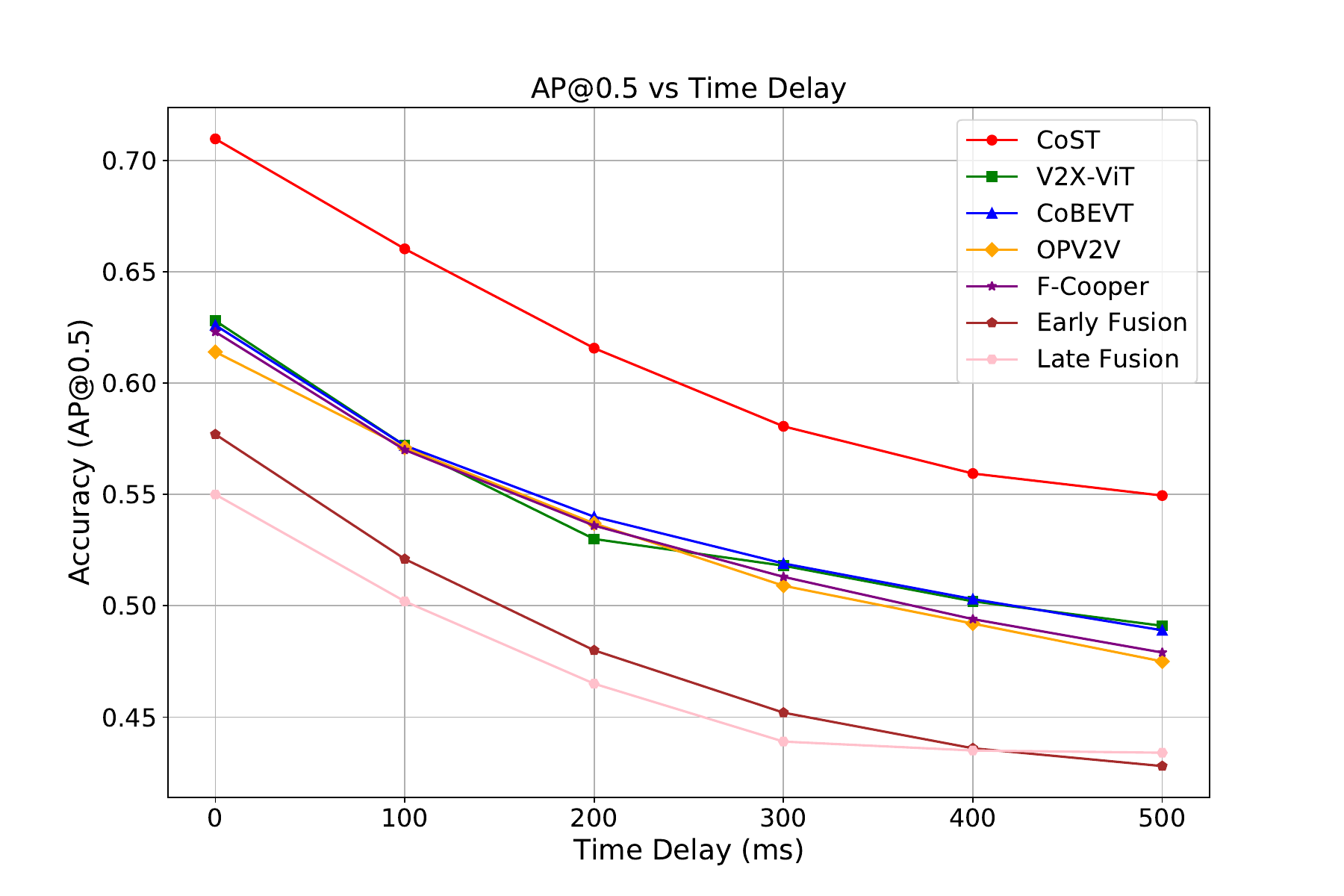} 
    }
    \vspace{-3mm}
    \caption{Robustness to localization errors and transmission latency. (a)(b)Initially trained in ideal conditions, models are tested under conditions with simulated localization errors by introducing Gaussian noise with a mean of $0$ to the LiDAR pose. The standard (Std) deviation is adjusted to vary the levels of noise for heading error (RYP dimensions) and positional error (XYZ dimensions). (c)Performance comparison of models under different times delays {0,100,200,300,400,500}ms.}
    \vspace{-4mm}
    \label{fig:positional_error}
\end{figure*}

\subsection{Performance-bandwidth Trade-off Analysis}
Reducing transmission bandwidth is a key challenge in Collaborative Perception. We calculate the communication volume as:
\begin{equation}
Comm = \log_2\left(\frac{N_c \times C \times 16}{8 \times 2^{20}}\right),
\end{equation}
where $N_c$ is the number of transmitted elements, $C$ is the number of channels, and data is transmitted in 16-bit floating-point format. This value, originally in bits, is converted to megabytes on a log2 scale.

Figure~\ref{fig:compress} illustrates the accuracy-bandwidth trade-off on V2V4Real. All methods utilizing PointPillars~\cite{lang2019pointpillars} generate identical BEV feature maps. Our CoST framework, particularly with the STT module, significantly reduces bandwidth (up to 1/100) with negligible accuracy loss. Omitting STT noticeably degrades performance. In practice, we control communication volume by thresholding $M_i^t$: varying the threshold within [0.0001, 1] achieves bandwidth compression from 1× to over 1000×, typically using 0.01 for around 60× compression.

The proposed STT and USTF modules can also be integrated into existing multi-agent fusion methods. In experiments with V2X-ViT~\cite{xu2022v2x} and CoBEVT~\cite{xu2022cobevt}, STT effectively reduces bandwidth usage. Integrating CoST substantially improves V2X-ViT's performance, but yields smaller gains for CoBEVT, likely because CoBEVT lacks mechanisms to handle time delays, thus limiting the benefit from CoST's temporal information.

\subsection{Robustness Analysis}

\textbf{Time Delay} in V2X communication degrades perception synchronization. We rigorously validate our model's robustness by injecting synthetic latencies (0-500ms) into collaborative feature streams  on V2V4Real dataset. Figure~\ref{fig:positional_error} shows that our method maintains superior accuracy (e.g., $66.03\%$ AP@0.7 under 100ms delays) across various time delay configurations.

\begin{figure}
    \includegraphics[width=1.\linewidth]{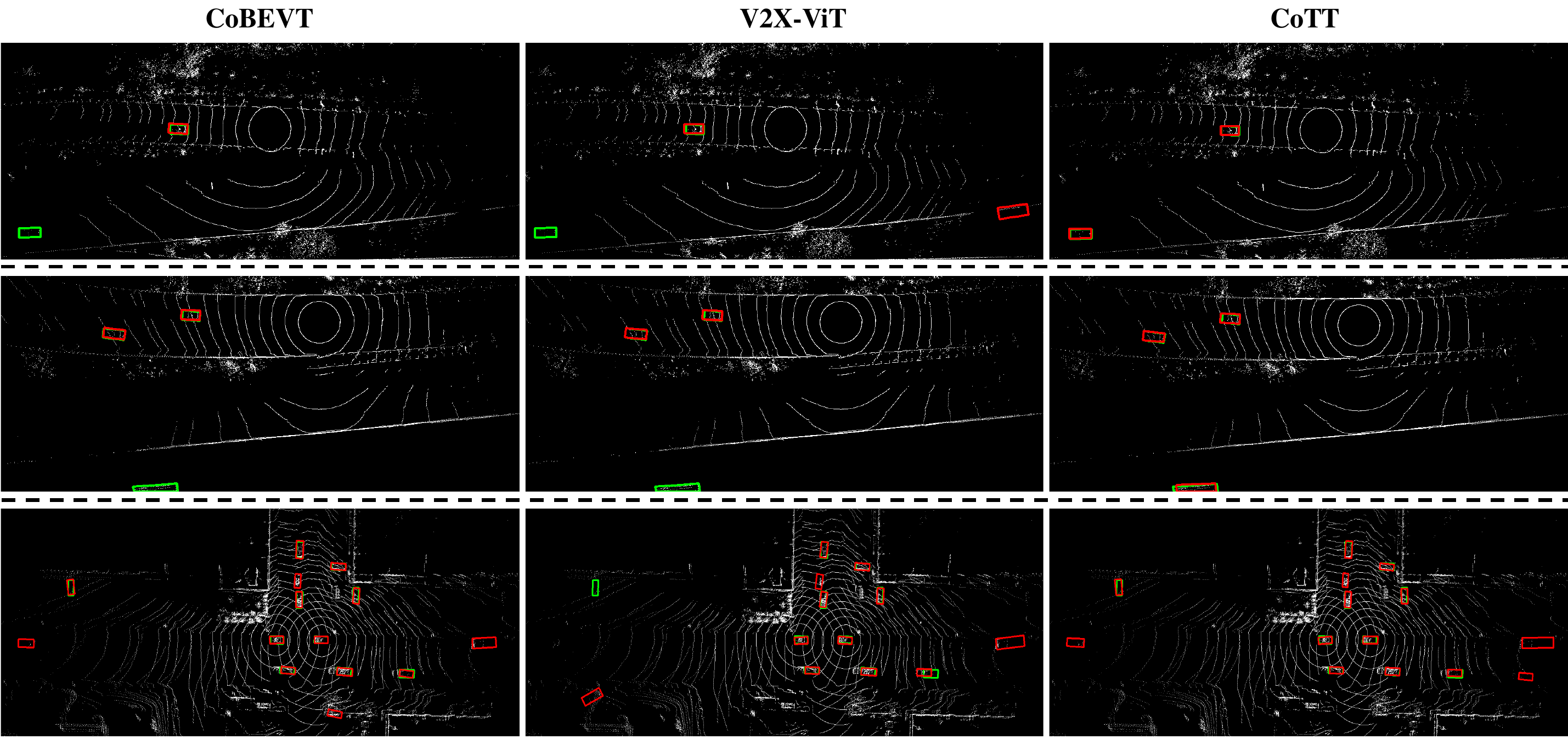}
    \vspace{-6mm}
    \caption{Qualitative comparison between sparse and dense scenes. The green and red 3D bounding boxes represent the ground truth and predictions. Our approach demonstrates superior accuracy in detection results. }
    \vspace{-4mm}
    \label{fig:vis}
\end{figure}

\noindent\textbf{Pose Error} is a significant issue in collaborative perception, encompassing both heading and localization errors.
Controlled experiments on V2XSet, as shown in Figure~\ref{fig:positional_error}, demonstrate that our method exhibits strong resilience to localization and heading errors, surpassing other methods notably in terms of AP@0.5.
More analysis about robustness appears in the supplementary material.

\subsection{Inference Time and Computational Cost}
Table \ref{tab:macs} summarizes the model size (Params), inference time (Speed), and performance (AP@0.5 and AP@0.7) of various collaborative perception methods on the V2V4Real dataset with single Tesla V100. Our CoST-S model, which excludes temporal information, outperforms previous methods with an inference time of 20.320 ms, a model size of 9.670M parameters, and AP scores of 67.68\% (AP@0.5) and 40.93\% (AP@0.7). When temporal information is incorporated, the CoST model shows only a slight increase in computational cost (22.908 ms and 9.790M parameters) while improving accuracy to 70.96\% (AP@0.5) and 43.97\% (AP@0.7). These results demonstrate that CoST effectively balances performance and computational efficiency.

\subsection{Qualitative Analysis}

Figure~\ref{fig:vis} shows detection results for CoBEVT, V2X-ViT, and CoST on both sparse and dense scenes in V2V4Real. Our model yields bounding boxes that closely match the ground truth, while competing methods exhibit noticeable errors. In sparse scenes (first and second rows), alternative approaches often miss distant objects and produce false positives, whereas our method maintains high accuracy. In dense scenes (last row), our approach achieves the most comprehensive detections, demonstrating its proficiency in effectively leveraging historical data.

\begin{table}[t]
\centering
\label{tab:macs}
\renewcommand{\arraystretch}{0.9}
\setlength{\tabcolsep}{1.2mm}{
\begin{tabular}{@{}cccccc@{}}
\toprule
Method & Speed(ms) & Params(M) & AP@0.5 & AP@0.7 \\ \midrule
\multicolumn{1}{c|}{V2X-ViT~\cite{xu2022v2x}}  & 169.787 & 12.462 & 0.6623 & 0.3651 \\
\multicolumn{1}{c|}{CoBEVT~\cite{xu2022cobevt}}  & 21.720 & 10.492 & 0.6560  & 0.3791 \\
\multicolumn{1}{c|}{CoST-S}  & 20.320 & 9.670 & 0.6768 & 0.4093 \\
\multicolumn{1}{c|}{CoST} & 22.908 & 9.790 & 0.7096 & 0.4397 \\ 
\bottomrule
\end{tabular}}
\vspace{-2mm}
\caption{Overview and Comparative Analysis of Key Features Across Various CP Techniques on the V2V4Real Dataset.}
\vspace{-5mm}
\label{tab:macs}
\end{table}

\section{Conclusion}
This paper introduces the Cooperative Spatio-temporal Transformer (CoST), a novel framework for collaborative perception in autonomous systems. CoST unifies historical and current agents into a shared spatio-temporal space by treating historical agents as time-delayed copies of current ones, enhancing feature fusion and reducing transmission bandwidth. The framework comprises three key components: Spatio-temporal Transmission (STT), Unified Spatio-temporal Fusion (USTF), and Multi-Agent Deformable Attention (MADA), which jointly optimize bandwidth usage and boost detection accuracy. Extensive experiments on V2V4Real, V2XSet, and DAIR-V2X demonstrate that CoST outperforms prior methods in both accuracy and efficiency. Moreover, STT and USTF can be seamlessly integrated with other frameworks to further improve performance while reducing communication overhead.

\vspace{2mm}
\noindent\textbf{Acknowledgements.} This research is supported in part by National Key R\&D Program of China (2022ZD0115502), National Natural Science Foundation of China (NO.62461160308, U23B2010), “Pioneer” and “Leading Goose” R\&D Program of Zhejiang (No. 2024C01161).

{
    \small
    \bibliographystyle{ieeenat_fullname}
    \bibliography{main}
}

\maketitlesupplementary 
\begin{figure*}[ht]

    \includegraphics[width=1\linewidth]{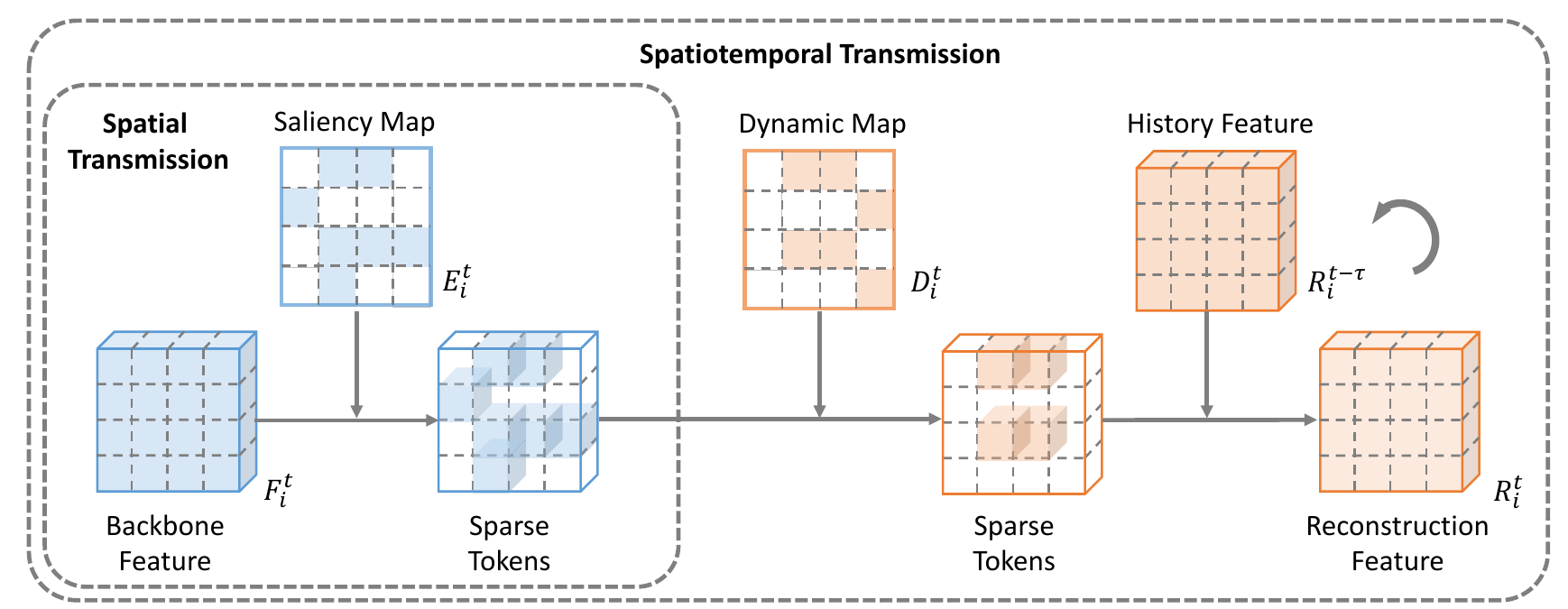}
    \caption{
    Our spatiotemporal transmission filters tokens based on saliency and motion, similar to selective attention in perception. These tokens are subsequently merged with historical features to reconstruct a comprehensive feature map. }
    
    \label{fig:fig3}
\end{figure*}
\section{Preliminary}
\label{sec:pre}
\subsection{Problem Formulation}

In a collaborative perception scenario involving $N$ agents, each agent $i$ owns unique observations $\{\mathcal{X}_{i}\}_{i=1}^N$ and the perception supervision $\{\mathcal{Y}_{i}\}_{i=1}^N$. The objective is to maximize the collective perception performance of all agents while ensuring that the transmission cost remains within a specified limit $G$. The process can be formulated as follows:
\begin{equation}
\begin{aligned}
\text{arg}\max_{\theta, \mathcal{M}} \sum_{i=1}^N g(\psi_{\theta}(\mathcal{X}_i, &\{\mathcal{M}_{i\rightarrow j}\}_{j=1}^N), \mathcal{Y}_i),\\ 
\text{s.t.} \sum_{i=1}^N\sum_{j=1,j\neq i}^N|\mathcal{M}_{i\rightarrow j}| &\leq G,
\end{aligned}
\end{equation} 
where $g(\cdot, \cdot)$ represents the perception evaluation metric, $\psi(\cdot)$ is the collaborative perception model with trainable  parameter $\theta$, and $\mathcal{M}_{i\rightarrow j}$ denotes the message transmitted from the $i$th agent to the $j$th agent.


\subsection{Collaboration Stages}
Collaborative perception in multi-agent systems involves several key stages to ensure comprehensive and accurate environmental understanding. Here, we outline the main stages of our proposed framework and their respective roles.

\noindent \textbf{V2X Metadata Sharing:}
The collaborative process begins with the sharing of meta information. Each agent shares metadata, including 6 degrees of freedom (6DoF) pose, extrinsic parameters, velocity, and agent type, which covers both infrastructure and vehicle type. One vehicle is designated as the ego agent. This shared meta information lays the foundation for subsequent feature extraction and fusion processes.

\noindent \textbf{Feature Extraction:}
Following metadata exchange, each agent extracts relevant features within its own view. We use the efficient PointPillar~\cite{lang2019pointpillars} for LiDAR point cloud feature extraction due to its minimal inference time and optimized memory utilization~\cite{xu2022opv2v}. PointPillar transforms the sparse point cloud into dense pillar tensors, which are then processed to obtain rich semantic BEV (Bird's Eye View) features $F_i^{t} \in \mathbb{R}^{H \times W \times C}$ at timestamp $t$ for agent $i$.

\noindent \textbf{Feature Communication:}
After extracting the features, agents exchange these features. Given the data-intensive nature of PointPillar features, reducing transmission bandwidth is crucial. Limited communication bandwidth in practical scenarios makes efficient feature communication a core challenge.
To address this, We introduce a spatiotemporal perspective where only dynamic object observations are transmitted, while static object observations are reused, reducing communication load. The ego agent combines selected tokens with historical features from the memory bank to form a reconstructed feature $R_i^t$ in $\mathbb{R}^{H \times W \times C}$, which then updates the memory bank for future use.

\noindent \textbf{Feature Fusion:}
Upon receiving features from other agents, the ego agent performs feature fusion. The goal is to integrate its own information with that received from other agents to derive the most comprehensive perceptual features. We employ a multi-agent fusion module that merges reconstructed features with ego features, producing a collaborative feature $B$ for the current timestamp in the space $\mathbb{R}^{H \times W \times C}$.

\noindent {\bf{Detection Head}}
In the final stage, the integrated perceptual features are fed into the detection head to predict the final perception results. This process involves two $1 \times 1$ convolution layers: one for box regression, outputting position, dimensions, and yaw angle of prediction boxes, and another for classification, generating a confidence score indicating the likelihood of each box containing an object or being background. The employed loss functions align with those of PointPillar~\cite{lang2019pointpillars}, including a smooth $L1$ loss~\cite{ren2016faster} for regression and focal loss~\cite{lin2017focal} for classification.

\section{Spatiotemporal Transmission}
\label{sec:stt}

As shown in Figure~\ref{fig:fig3}, the STT module in our CoST framework reduces communication bandwidth by transmitting only dynamic object features instead of the full BEV map. Conventional methods reduce feature channels~\cite{xu2022v2x,xu2023v2v4real}, but they neglect the temporal redundancy where static information remains unchanged over time. While Where2comm~\cite{huwhere2comm} transmits sparse tokens from key object regions, it does not leverage the fact that static regions can be reconstructed from historical data. In our approach, we focus on dynamic features that change over time, thereby avoiding the re-transmission of redundant static information while ensuring a complete scene representation.

\begin{figure}[t]
\vspace{-4mm}
\centering
\includegraphics[width=0.95\linewidth]{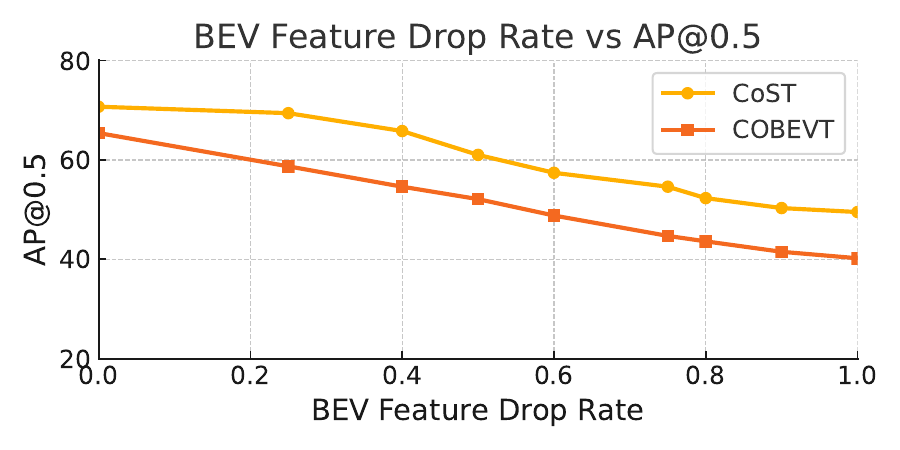}
\vspace{-3mm}
\caption{
Performance under randomly dropped transmitted features on V2V4Real. CoST shows better robustness with minimal accuracy drop.
}
\label{fig:comm_loss}
\vspace{-2mm}
\end{figure}

\textbf{Robustness to Communication Loss.}
To further evaluate the robustness of our framework under lossy communication, we simulate perturbations by randomly dropping a portion of transmitted feature tokens. As shown in Figure~\ref{fig:comm_loss}, CoST exhibits stronger resilience compared to baselines, with significantly smaller degradation in detection accuracy. This robustness stems from our use of temporal memory to  mitigate the impact of missing data during transmission.

\section{Training Details}
\label{sec:detail}
All detection models utilize PointPillar~\cite{lang2019pointpillars} to extract the BEV features from the point cloud and the feature channel $C$ is set as 256. The models are trained over 60 epochs with a batch size of 4 per GPU (Tesla V100), a learning rate set at 0.001, and incorporating learning rate decay using a cosine annealing strategy~\cite{loshchilov2017decoupled}.
Regarding the USTF module, during training, we use data samples of three consecutive frames with a time interval $\tau$ set to 2 frames to obtain more context information. During testing, we process the frames sequentially from the first frame to the last. As for the STT compression module, it undergoes an additional two-stage training process with the other pre-trained modules frozen.
To ensure consistency in evaluation, standard settings are followed on the V2V4Real~\cite{xu2023v2v4real} dataset, including typical data augmentations for point cloud data. Conversely, no data augmentations are applied to the V2XSet~\cite{xu2022v2x} and DAIR-V2X~\cite{yu2022dair} datasets. The model optimization is performed using AdamW~\cite{kingma2014adam} with a weight decay of $1\times10^{-2}$ to fine-tune the models. 

\section{Datasets}
\label{sec:dataset}
\noindent {\bf{V2V4Real:}}
V2VReal is a large-scale real-world V2V dataset collected by two vehicles with multi-modal sensors navigating through diverse scenarios. Covering a driving area of 410 km, V2V4Real includes $20K$ LiDAR frames, $240K$ annotated 3D bounding boxes across $5$ classes, and HDMaps that encompass all driving routes.

\noindent {\bf{V2XSet:}}
V2XSet integrates V2X collaboration with realistic noise simulation. The dataset comprises a total of $11,447$ frames from CARLA~\cite{dosovitskiy2017carla} and OpenCDA~\cite{xu2021opencda}, split into training, validation, and test sets with $6,694$, $1,920$, and $2,833$ frames, respectively. It includes five types of roadways: straight segments, curvy segments, midblocks, entrance ramps, and intersections. Each scene features between two to seven intelligent agents for collaborative perception.

\noindent {\bf{DAIR-V2X:}}
DAIR-V2X is a substantial real-world V2I dataset without V2V collaboration. It comprises three parts: DAIR-V2X-C, DAIR-V2X-I, and DAIR-V2XV. Notably, DAIR-V2X-C contains sensor data from vehicles and infrastructure, with $38,845$ frames each from cameras and LiDAR, and around 464,000 3D bounding boxes categorized into 10 distinct classes.
\section{Qualitative Analysis}
\begin{figure*}[t]
    \begin{center}
    \includegraphics[width=1.\linewidth]{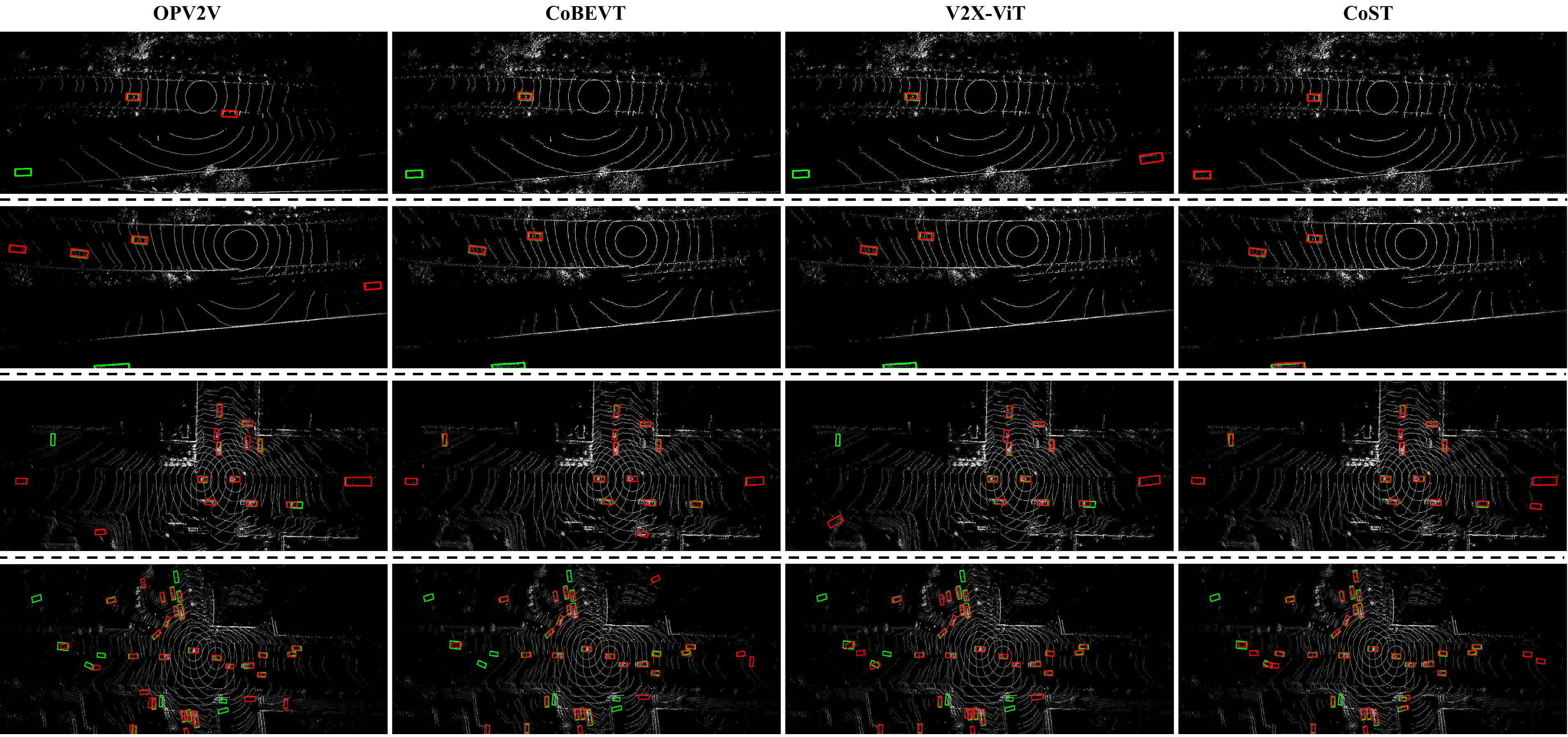}
    \end{center}
    \caption{Qualitative comparison between sparse and dense scenes. The green and red 3D bounding boxes represent the ground truth and predictions, respectively. Our approach demonstrates superior accuracy in detection results. }
    \label{fig:vis_4}
\end{figure*}
Figure~\ref{fig:vis_4} illustrates the detection visualization for OPV2V, CoBEVT, V2X-ViT, and CoTT across sparse and dense scenes in V2V4Real. Our model produces precise bounding boxes that closely align with the actual ground truth, contrasting with competing methods that exhibit notable discrepancies.
Specifically, in sparse scenes with fewer than three ground truths, observed in the initial and second rows, alternative methods frequently overlook distant ground truth boxes and may produce false positives. Conversely, our approach delivers accurate predictions in such scenarios.
Moreover, our methodology achieves the most thorough detection outcomes in dense environments represented in the last two rows of Figure~\ref{fig:vis_4}, demonstrating its proficiency in effectively leveraging historical data.

\begin{figure}[]
    \begin{center}
    \includegraphics[width=1.\linewidth]{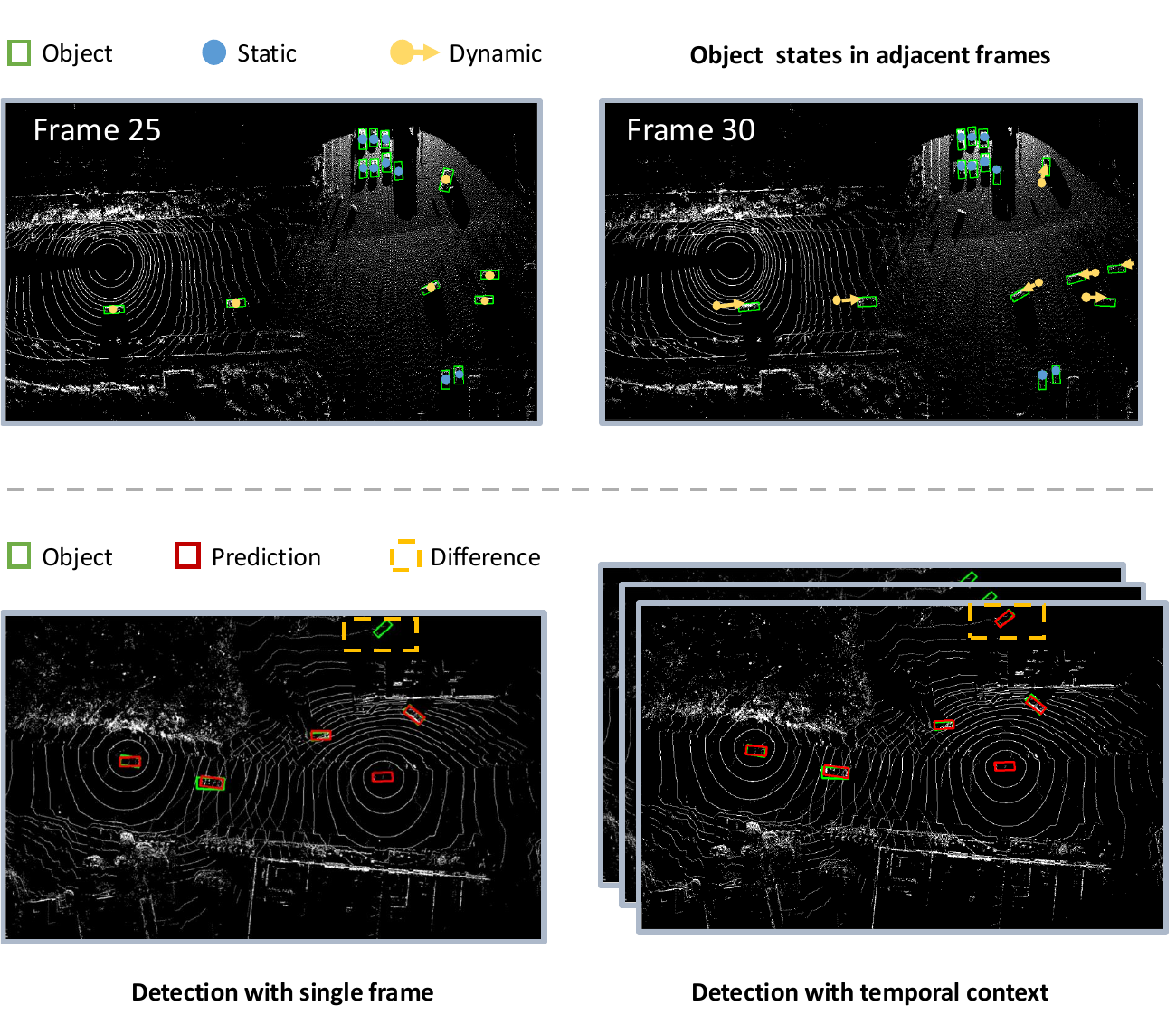}
    \end{center}
    \caption{Visualization for object states on adjacent frames and comparing detection with single frame and temporal context.}
    \label{fig:adjacent}
\end{figure}

We visualize the ground truth objects/boxes in adjacent frames in Figure \ref{fig:adjacent}.
Only the six cars positioned in the center of the frame have moved between frame $25$ and frame $30$, while the state of the other vehicles has remained unchanged.
Besides, the temporal context benefits the detection result and reduces the omission of detection results, as shown in the bottom of Figure \ref{fig:adjacent}.

\section{Robustness Analysis}
\label{sec:robustness}
\textbf{Time Delay} poses a notable challenge in real-world V2X communications, leading to a lack of synchronization between the ego vehicle's functionalities and information received from collaborative agents. It is essential for collaborative perception techniques to effectively handle such time delays. In this research, we investigate the resilience of our model towards time delays. We enhanced our approach by utilizing a model originally trained in an ideal environment, introducing latencies uniformly distributed between 0 to 500 milliseconds. To assess our model, we conducted tests with fixed latencies.
In scenarios featuring a fixed delay (as depicted in Figure \ref{fig:positional_error} in the main text), a consistent latency of up to 500 milliseconds was added to the feature transmission from each agent to the ego vehicle.
Our findings reveal that our method demonstrates substantial resilience to latency within V2V4Real.
In contrast, both early fusion and late fusion methodologies exhibit a notable decline in performance when confronted with latency issues, indicating a lack of robustness.
Consistently, our method surpasses other techniques in accuracy across various time delay configurations.
Even in settings with noisy 100 milliseconds time delays, our method achieves an accuracy of $66.03\%$ AP@0.7, outperforming other cooperative approaches under optimal conditions.

\noindent\textbf{Pose Error} is a significant issue in collaborative perception, encompassing both heading error and localization error. In our experiments, we selected the V2XSet dataset for validation because the real-world database V2V4Real already contains inherent pose error noise, making the effects of manually added noise less discernible.
To mimic real-world inaccuracies in our experiments, we introduced localization noise without modifying the models' settings. This noise follows a Gaussian Distribution with a mean of zero and an adjustable standard deviation, emphasizing the importance of robustness in collaborative perception models against localization inaccuracies. The results from the V2XSet dataset shown in Figure \ref{fig:positional_error} in the main text indicate that heading errors have a more significant impact on performance compared to positional errors. It is evident that introducing localization errors of $0.1$ and $0.2$ has minimal impact on the accuracy of our method. Notably, our method exhibits strong resilience to localization and heading errors, surpassing other methods notably in terms of AP@0.5.

\end{document}